\pdfoutput=1

\documentclass[11pt]{article}

\usepackage[]{acl}

\usepackage{times}
\usepackage{latexsym}

\usepackage[T1]{fontenc}

\usepackage[utf8]{inputenc}

\usepackage{microtype}

\usepackage{inconsolata}

\usepackage{array}
\usepackage{booktabs}
\usepackage{multirow}
\usepackage{amsmath}
\usepackage{amssymb}
\usepackage{graphicx}
\usepackage{enumerate}
\usepackage{diagbox}
\usepackage{enumitem}
\usepackage{color}
\usepackage{xcolor}
\usepackage{soul}

\newcommand{\ourmethod}{ConstraintChecker} 

\title{ConstraintChecker: A Plugin for Large Language Models\\ to Reason on Commonsense Knowledge Bases}

\author{Quyet V. Do, Tianqing Fang, Shizhe Diao, Zhaowei Wang, Yangqiu Song\\
Department of Computer Science and Engineering, HKUST, Hong Kong SAR, China \\
\texttt{\{vqdo, tfangaa, sdiaoaa, zwanggy, yqsong\}@cse.ust.hk}
}

\begin{document}
\maketitle

\begin{abstract}
Reasoning over Commonsense Knowledge Bases (CSKB), i.e., CSKB reasoning, has been explored as a way to acquire new commonsense knowledge based on reference knowledge in the original CSKBs and external prior knowledge.
Despite the advancement of Large Language Models (LLM) and prompt engineering techniques in various reasoning tasks, they still struggle to deal with CSKB reasoning.
One of the problems is that it is hard for them to acquire explicit relational constraints in CSKBs from only in-context exemplars, due to a lack of symbolic reasoning capabilities ~\cite{bengio2021deeplearning}.
To this end, we proposed \textbf{\ourmethod{}}, a plugin over prompting techniques to provide and check explicit constraints.
When considering a new knowledge instance, \ourmethod{} employs a rule-based module to produce a list of constraints, then it uses a zero-shot learning module to check whether this knowledge instance satisfies all constraints. 
The acquired constraint-checking result is then aggregated with the output of the main prompting technique to produce the final output.
Experimental results on CSKB Reasoning benchmarks demonstrate the effectiveness of our method by bringing consistent improvements over all prompting methods. Codes and data are available at \url{https://github.com/HKUST-KnowComp/ConstraintChecker}.
\end{abstract}


\section{Introduction}

\begin{figure}[t]
    \centering
    \includegraphics[width=1\linewidth]{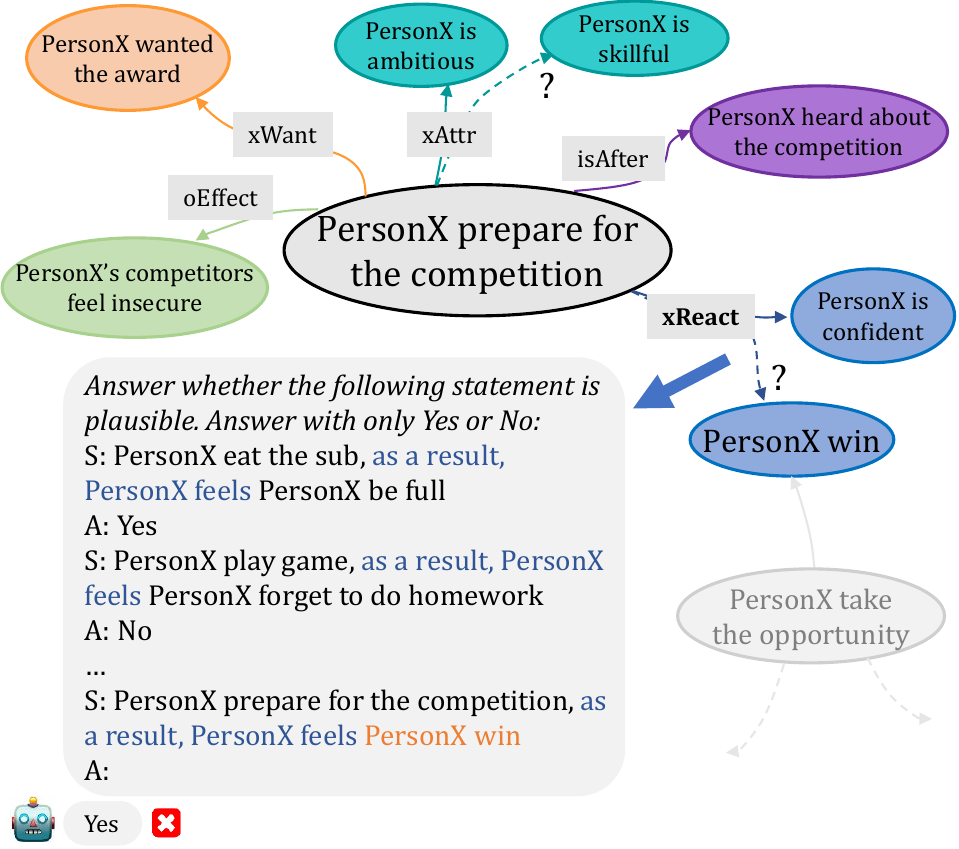}
    \caption{Examples of CSKB Reasoning. Solid arrows represent existing triples (i.e., instances) in CSKBs, while the dashed arrows with question marks represent new knowledge triples which will be determined if they are commonsense. LLMs often fail to acquire the explicit relational constraints in CSKBs, hence making wrong predictions for many new knowledge triples.}
    \label{fig:problem}
    \vspace{-0.1in}
\end{figure}

Commonsense Knowledge Bases (CSKB) Reasoning, as one of many commonsense reasoning tasks, has been well explored in Natural Language Processing for the past few years.
As human-annotated CSKBs~\cite{singh2017conceptnet, sap2019atomic, mostafazadeh2020glucose} are usually incomplete and of a small coverage, reasoning over CSKBs, i.e., CSKB reasoning, is a way for expansion. CSKB reasoning is defined as determining whether a new knowledge triple \textit{(head event, relation, tail event)} is commonsense (in other expressions, being plausible or having label 1) based on the reference knowledge in original CSKBs as well as external prior knowledge~\cite{fang2023ckbpv2, davison2019commonsense}.
Expanding CSKBs via such a reasoning process can lead to better and broader commonsense knowledge as valuable resources to augment AI models in various aspects, such as visual reasoning ~\cite{zellers2019vcr}, text generation~\cite{zhou2021commonsense,ilievski2021story}, or building more capable knowledge models for further downstream applications ~\cite{yu2022cocolm,hwang2021comet,wang-etal-2023-cat}. 

Recently, inspired by the emergence of Large Language Models (LLMs) that can perform well in many commonsense reasoning tasks~\cite{qin2023chatgpt, bian2023chatgpt}, ~\citet{chan2023chatgpt} attempted to use LLMs for a CSKB Reasoning benchmark named CSKB Population (CKBP)~\cite{fang2023ckbpv2}. However, the result shows that LLMs still fall short in the benchmark, even with a large number of in-context examples.
One of the problems is that LLMs find it hard to acquire the explicit relational constraints in CSKBs, hence making wrong predictions.
In the example in Figure ~\ref{fig:problem}, the \texttt{xReact}\footnote{By definition in ~\cite{sap2019atomic}, a tail event of the \texttt{xReact} relation expresses how the protagonist feels after the corresponding head event.} relation in CSKBs requires the tail event of the knowledge triple to express a mental state, such as ``\textit{PersonX} is confident'', instead of an action, such as ``\textit{PersonX} win''.
Meanwhile, LLMs fail to recognize the constraint from in-context exemplars, thus making the judgment mainly based on the semantics of the head and tail events.
It leads to an incorrect prediction that the triple (\textit{PersonX} prepare for the competition, \texttt{xReact}, \textit{PersonX} win) is plausible. 
In light of this, many advanced prompting techniques, such as Chain-of-Thouht (CoT) \cite{wei2022cot}, Least-to-Most \cite{zhou2023leasttomost}, Active-CoT \cite{diao2023active}, etc., can be possible alternatives for improvements. Nonetheless, they are task-agnostic and suffer from the inherent shortcoming of LLMs in inducing the rules in CSKBs (which we refer as symbolic reasoning ability),
as current deep learning still struggles to deal with symbolic and high-level concepts reasoning tasks ~\cite{bengio2021deeplearning,huang-chang-2023-towards,
pan2023logiclm}.

To this end, we propose \textbf{\ourmethod{}}, \textbf{a plugin component} for LLMs to handle the problem of explicit constraints in CSKB reasoning.
\ourmethod{}
supports LLMs' reasoning as an independent component in addition to the \textbf{main-task component} that determines whether a knowledge triple is commonsense or not.
There are two modules in this plugin.
Given a knowledge triple \textit{(head event, relation, tail event)}, we first employ a rule-based/symbolic module to produce a list of constraints based on the relation.
The list is then passed to a zero-shot learning module, where we construct constraint-checking questions and use the same LLM as in the main-task component in a zero-shot manner to check whether the instance satisfies all constraints.
The acquired constraint-checking result is then aggregated with the prediction from the main-task component by logical conjunction to produce the final prediction.

We implement \ourmethod{} and conduct extensive experiments on a CSKB Reasoning benchmark CKBPv2 ~\cite{fang2023ckbpv2} as well as a synthetic discriminative version of ATOMIC$_{20}^{20}$ (in short, SD-ATOMIC$_{20}^{20}$), over two large language models: ChatGPT (gpt-3.5-turbo-0301) and GPT3.5 (text-davinci-003).
On both language models, \ourmethod{} improves over prompting techniques (as the main-task component) by a significant margin in different metrics, achieving the best result on both the benchmarks CKBPv2 and SD-ATOMIC$_{20}^{20}$.
Further analyses and ablation studies show the effect of each of the considered constraints as well as different choices of prompt design in \ourmethod{}, and the superiority of its plug-and-play design over the single-prompt counterpart.

To summarize, our contribution is \textbf{two-fold}: 
(1) We propose \ourmethod{}, an independent plugin that handles the problem of explicit constraints in CSKB reasoning to improve over main-task prompt methods, and
(2) We conduct extensive experiments on two CSKB Reasoning benchmarks CKBPv2 and SD-ATOMIC$_{20}^{20}$, to demonstrate our method's effectiveness and advantages over other advanced prompting techniques.

\section{Background and Related Works}
\label{sec:related_works}

\subsection{CSKB Reasoning}
Commonsense knowledge bases store commonsense knowledge in the format of \textit{(head event, relation, tail event)} triples. Reasoning on CSKBs includes two main settings, a discriminative and a generative one. They are formally defined as: Given $B = \{{(h,r,t)|h \in H, r \in R, t \in T}\}$ (where $H/R/T$ is the set of head events/relations/tail events) the reference knowledge base, the discriminative setting ~\cite{DBLP:conf/www/FangZWSH21, fang-etal-2021-benchmarking,fang2023ckbpv2} assigns a binary label $y \in \{0,1\}$ to a new knowledge triple $T = (h,r,t)$ to indicate whether the triple $T$ is commonsense or not; while the generative setting ~\cite{bosselut-etal-2019-comet,hwang2021comet} generates new commonsense knowledge triples $T' = (h,r,t')$ based on existing head events $h$ and relations $r$. Although our method can be adapted to the generative setting, the evaluation of our method is more complex than that in the discriminative setting. Therefore, in this work, we only focus on the discriminative setting.


In terms of benchmarks with the discriminative setting, to our best knowledge, CKBPv2~\cite{fang2023ckbpv2} and its predecessor CKBPv1~\cite{fang-etal-2021-benchmarking} stand as only comprehensive CSKB Reasoning benchmarks, which cover the knowledge on four popular CSKBs (ConceptNet; \citealp{singh2017conceptnet}, ATOMIC; \citealp{sap2019atomic}, ATOMIC$_{20}^{20}$; \citealp{hwang2021comet}, and GLUCOSE; \citealp{mostafazadeh2020glucose}). Nonetheless, since the reference CSKBs are popular and widely used among the NLP community, we believe CKBPv2 is representative enough in terms of CSKB reasoning. Besides, to ensure the reliability of our result in this work, we synthesize a discriminative-setting dataset from ATOMIC$_{20}^{20}$ which is designated for the generative setting. Indeed, the two benchmark datasets inherit similar head/tail events' formats and the same relation list from ATOMIC$_{20}^{20}$.


Meanwhile, in term of methodology, despite previous efforts to CSKB reasoning, most of them are based on knowledge base embeddings~\cite{li-etal-2016-commonsense, Malaviya2019ExploitingSA, hua-zhang-2022-system} or (lightweight) fine-tuning pre-trained language models~\cite{yao2019kg, fang-etal-2021-benchmarking, zhang2023making}, and less effort has been dedicated to studying how to use LLMs for CSKB reasoning via prompting. We address this research gap by studying a constraint-checking plugin to enhance the performance of LLMs.

\subsection{Constraint Modelling in Traditional Knowledge Bases}
Integrating rules or constraints into reasoning systems on traditional knowledge bases (KB) and knowledge graphs (KG) has long been studied~\cite{Wang2015KnowledgeBC, denis2015typeconstraint, ding-etal-2018-improving, zhang2019iterE, lan2023knowledge}. While ~\citet{Wang2015KnowledgeBC} aimed to incorporate rules seamlessly into embedding models for KB completion during inference time by formulating inference as an integer linear programming (ILP) problem, ~\citet{denis2015typeconstraint} studied the effect of type-constraints on the statistical modeling with latent variable models for large knowledge graphs. In a more recent time, other works such as ~\citet{ding-etal-2018-improving, zhang2019iterE, lan2023knowledge} attempt to improve KG embedding by modelling rules and constraints in the learning objective. Our work, by contrast, introduces a novel augmentation paradigm which employs an explicit use of constraints in the inference time to improve the performance of large language models on CSKB reasoning.

\subsection{Prompting Methods in LLMs}
While simple prompt engineering and vanilla in-context learning have already witnessed a remarkable performance in terms of various NLP tasks, there are more sophisticated prompt paradigms to elicit better reasoning capabilities.
One representative paradigm is chain-of-thought (CoT) prompting~\cite{wei2022cot}, which enrichs the few-shot examples with explicit reasoning steps towards the final answers, leading to the emergence of many complex reasoning abilities such as arithmetic and commonsense reasoning.
Following CoT, other techniques adopt self-consistency~\cite{wang2023selfconsistency}, least-to-most that break down each question to sub-steps~\cite{zhou2023leasttomost}, pre-trained verifier to validate the reasoning path~\cite{li2023making}, diversity-based method for CoT selection~\cite{zhang2022automatic}, 
restrict explicit and rigorous deductive reasoning of intermediate CoT reasoning process~\cite{ling2023deductive},
uncertainty-based method for CoT selection and annotation~\cite{diao2023active}, and automatic prompt augmentation and selection with CoT~\citep{shum2023automatic}.
Our work differs from those CoT-based prompt techniques in that we study add-on constraints to be applied to the result of any prompting technique.

\begin{figure*}[t]
    \centering
    \includegraphics[width=1\linewidth]{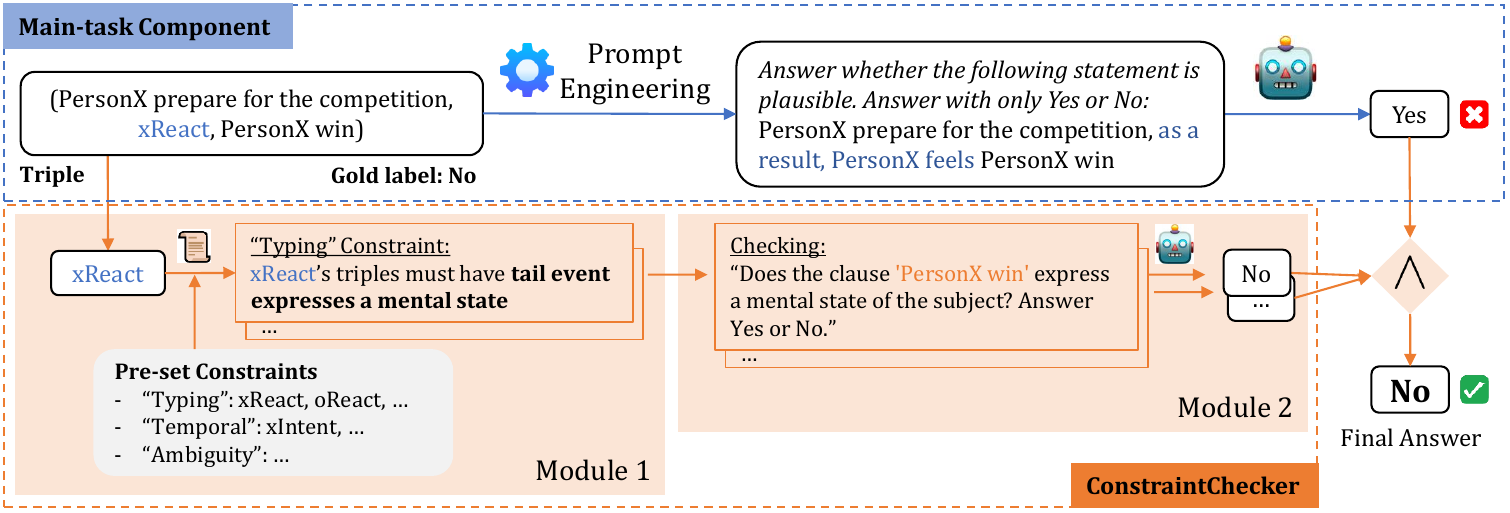}
    \vspace{-0.25in}
    \caption{Illustration of \ourmethod{}.
    For each instance, Module 1 queries \textit{preset rules} to get all relational constraints corresponding to the instance's relation.
    Module 2 then constructs questions accordingly to ask whether each constraint is satisfied, and passes these questions to the backbone LLM to get predictions.
    Together with the prediction from the main-task component, we use the logical conjunction (i.e., $\wedge$ or \texttt{AND} operator) to aggregate the final prediction.
    Note that ``Prompt Engineering'' refers to baseline prompting methods (subsection ~\ref{sec:setup}).
    }
    \label{fig:method_illustration}
    \vspace{-0.1in}
\end{figure*}

\section{\ourmethod{}} 

An overview of our proposed \ourmethod{} is shown in Figure ~\ref{fig:method_illustration}.
The CSKB reasoning task we focus on is inherently a binary classification task and the expected outputs are either \textit{plausible} or \textit{implausible}.
\ourmethod{} consists of two modules, entitled Module 1 and Module 2.
For each instance, Module 1 queries \textit{preset rules} to get all relational constraints corresponding to the instance's relation.
Module 2 then constructs questions accordingly to ask whether each constraint is satisfied, and passes these questions to the backbone LLM to get predictions.
Together with the prediction from the main-task component, we use the logical conjunction (\texttt{AND} operator) to aggregate the final prediction.
In fact, \ourmethod{} only has the effect on instances that are predicted as commonsense (or ``Yes'', corresponding to \textit{plausible}) by the main-task component, and can only change the prediction from ``Yes'' to ``No'', in view of the nature of logical conjunction. Thus, it targets and corrects False-Positive predictions.

In this section, we elaborate on how we select the pool of constraints and the preset rules to map relations to constraints in Module 1, as well as the constraint-checking prompt design in Module 2 concerning the two benchmarks.


\subsection{Constraints Selection}
We follow the definitions of CSKB relations in previous works, including the taxonomy of \textit{if-then} reasoning types in \citet{sap2019atomic} and human-readable templates for crowdsourced evaluation in \citet{hwang2021comet}\footnote{The templates are shown in Table ~\ref{tab:relation_taxonomy} and ~\ref{tab:readable_template} in the Appendix.}, to derive the set of considered constraints and the rule to apply constraints.
For example, the readable template ``as a result, \textit{PersonX} feels'' of the \texttt{xReact} relation suggests the ``temporal'' constraint, in which the head event must happen before the tail event, and the taxonomy of \texttt{xReact} implies the ``typing'' constraint, in which the tail event must express a mental state.
Note that the template ``as a result, \textit{PersonX} feels''  of \texttt{xReact} may not strictly impose the typing constraint on the tail event, due to a subtle problem in natural language.
For example, for human, two text sequences ``as a result, \textit{PersonX} feels \textit{PersonX} will win'' and ``as a result, \textit{PersonX} feels \textit{PersonX} is confident'' all make sense, although ``\textit{PersonX} will win'' completely does not express a mental state of \textit{PersonX}.

In addition, as ~\citet{davis2023survey} suggests that many commonsense datasets have significant portions of ambiguous instances, we also consider the ``ambiguity'' constraint.

Among of possible constraints, we shortlisted the most likely needed constraints, namely typing, temporal, and ambiguity constraints. The formal definition of each constraint is as follows:
\begin{itemize}[leftmargin=*,label=$\bullet$,noitemsep,partopsep=0pt,topsep=0pt,parsep=0pt]
\item \textbf{Typing:}  The tail event has to express the type of content (one among three types: activity, mental state, persona) that the relation expects. For example, \texttt{xReact}'s tail events need to express a mental state, while \texttt{xAttr}'s tail events need to express a persona. 
\item \textbf{Temporal:} The (estimated) temporal order (i.e. before or after) of the head event and the tail event must follow the order derived from the definition/human-readable template of the relation. For example, for \texttt{HinderedBy} relation, the head event must happen after the tail event.
\item \textbf{Ambiguity:} The meaning of the head and tail events must be grammatically complete and semantically informative. For example, ``\textit{PersonX} order a salad'' is not ambiguous, while ``\textit{PersonX} would like'' is ambiguous. 
\end{itemize}


\begin{table*}[t]
\small
\centering
\setlength{\tabcolsep}{4.5pt} 
\renewcommand{\arraystretch}{1} 
\begin{tabular}{c|l|l}

\toprule
Constraint & Relations on effect (pre-pilot-study) & Relations on effect (post-pilot-study) \\
\midrule
Typing & \texttt{xReact (m), oReact (m), xAttr (p)} & \texttt{xReact (m), oReact (m), xAttr (p)} \\
\midrule
Temporal & \texttt{xIntent (a), xNeed (a), Causes (b), HinderedBy (a)} & \texttt{xIntent (a), xNeed (a)} \\
\midrule
Ambiguity & All relations & No relation \\
\bottomrule

\end{tabular}
\caption{Relations on range of effect of each constraint. For Typing constraint, the notation \texttt{(m)} and \texttt{(p)} denote the required type ``mental state'' and ``person'' of the tail event. For Temporal constraint, the notation \texttt{(a)} and \texttt{(b)} denote the required estimated order ``after'' and ``before'' of the head and tail events.}
\label{tab:constraint_rule}
\vspace{-0.1in}
\end{table*}

\subsection{Preset Rules}
Each relation will be mapped into a set of constraints based on the aforementioned taxonomy and templates, as well as human-readable templates used by the main-task component in terms of how well the template of the relation semantically reflects the constraints of that relation.
For example, the template of \texttt{xReact} ``as a result, \textit{PersonX} feels'' does contain the phrase ``as a result'' representing the temporal constraint which is needed to check.
To refine the rule, we conduct a pilot study with ChatGPT on the development set of CKBPv2 to estimate the effectiveness of designated constraints on each relation.
According to the results of the pilot study (Appendix \ref{sec:pilot}), we remove ineffective constraint-relation pairs to refine the rule, as shown in Table ~\ref{tab:constraint_rule}.
In fact, the ineffectiveness of Ambiguity constraint suggests that \textit{randomly taking a constraint and then cherry-picking its effective constraint-relation pairs would not work}, instead we need to derive the rules from prior knowledge about relations. 
We further conduct an ablation study on the main experiments w.r.t. ChatGPT on CKBPv2 to show the ineffectiveness of removed relation-constraint pairs (Section \ref{sec:ablation}).

\subsection{Constraint-Checking Prompt Design}
As we use a zero-shot LLM to check constraints, we construct questions for derived constraints in a direct question-answering manner.
For example, for the typing constraint, which requires the tail event of the triple to express a mental state, we design a prompt as ``Does the clause \textit{<tail event>} express a mental state of the subject? Answer Yes or No''. 
Thanks to the robustness of LLMs and the fact that constraint satisfaction is a relatively simple task that does not require complex reasoning, exemplars for constraint-checking questions are not needed.
For each constraint, we design two templates to seek the best one. 
We provide an analysis of different prompt choices later in Section \ref{sec:ablation}.
Overall, we choose the following prompt designs for typing and temporal constraints respectively.\\
\textbf{Typing:} ``Which aspect (among three options 1. event/activity, 2. persona, 3. mental state) of the subject does the clause \textit{<tail event>} express. Answer the choice only.'' \\
\textbf{Temporal:} ``Which one of the following two statements is more plausible:
0. \textit{<tail event>} before \textit{<head event>},
1. \textit{<tail event>} after \textit{<head event>}.
Answer 0 or 1 only.''

Since the chosen prompts do not standardly question whether the constraint is satisfied, 
we use a snippet of code to convert the acquired prediction into the Yes/No answer for the standard constraint-checking question.


\vspace{-0.1in}
\section{Experiments}


\subsection{Benchmarks}


We use CKBPv2~\cite{fang2023ckbpv2} and SD-ATOMIC$_{20}^{20}$ as the CSKB reasoning benchmarks for evaluation. For CKBPv2, to reduce the computational cost while keeping the same data distribution w.r.t two attributes \textit{relation} and \textit{label}, we use stratified sampling to down-scale the test set, hence forming a set of 979 instances. Meanwhile, SD-ATOMIC$_{20}^{20}$ is curated from the test set of ATOMIC$_{20}^{20}$ using random relation/tail event replacement as the negative sampling strategy, resulting in a set of 1000 instances with the equal numbers of positive and negative instances. More details about the two benchmarks can be found in Appendix \ref{sec:benchmark}.

\begin{table*}[t]
\small
\centering
\setlength{\tabcolsep}{4.5pt} 
\renewcommand{\arraystretch}{1.1} 
\begin{tabular}{l|ll|ll|ll|ll}
\toprule
\multirow{3}{*}{Method} & \multicolumn{4}{c|}{CKBPv2} & \multicolumn{4}{c}{SD-ATOMIC$_{20}^{20}$} \\
\cline{2-9}
& \multicolumn{2}{c|}{ChatGPT} & \multicolumn{2}{c|}{GPT3.5} & \multicolumn{2}{c|}{ChatGPT} & \multicolumn{2}{c}{GPT3.5} \\
\cline{2-9}
& Acc & F1 & Acc & F1 & Acc & F1 & Acc & F1 \\

\midrule
Zero-shot & 67.58 & 47.48  &  73.92 & 37.16    &    59.6 & 61.3  &  59.7 & 62.72  \\
~ + \ourmethod{} & 69.19 & 48.45 \scriptsize{(+0.97)} &  74.97 & 38.09 \scriptsize{(+0.93)}    &    63.4 & 63.62 \scriptsize{(+2.32)} &  62.2 & 64.2 \scriptsize{(+1.48)} \\
\midrule
Few-shot (Random) & 69.94 & 48.84  &  72.12 & 49.76    &    59.4 & 60.43  &  57.4 & 61.13  \\
~ + \ourmethod{} & 71.67 & 50.15 \scriptsize{(+1.31)} &  73.82 & \textbf{51.07} \scriptsize{(+1.31)}    &    62.0 & 62.0 \scriptsize{(+1.57)} &  60.2 & 62.73 \scriptsize{(+1.6)} \\
\midrule
Few-shot (KATE) & 68.0 & 44.94  &  69.87 & 47.41    &    61.3 & 60.55  &  58.5 & 61.4  \\
~ + \ourmethod{} & 69.73 & 46.01 \scriptsize{(+1.07)} &  71.81 & 48.89 \scriptsize{(+1.48)}    &    63.8 & 62.13 \scriptsize{(+1.58)} &  61.7 & 63.28 \scriptsize{(+1.88)} \\
\midrule
Few-shot (KATE-s) & 67.35 & 45.99  &  69.25 & 47.52    &    59.0 & 59.08  &  60.3 & 62.51  \\
~ + \ourmethod{} & 69.19 & 46.95 \scriptsize{(+0.96)} &  71.09 & 48.68 \scriptsize{(+1.16)}    &    61.7 & 60.72 \scriptsize{(+1.64)} &  63.0 & 64.15 \scriptsize{(+1.64)} \\
\midrule
Zero-shot-CoT & 62.14 & 42.37  &  77.25 & 39.49    &    49.9 & 50.25  &  58.4 & 57.89  \\
~ + \ourmethod{} & 64.35 & 43.58 \scriptsize{(+1.21)} &  77.93 & 40.21 \scriptsize{(+0.72)}    &    52.7 & 51.69 \scriptsize{(+1.44)} &  60.8 & 59.34 \scriptsize{(+1.45)} \\
\midrule
Few-shot-CoT & 76.92 & 48.41  &  62.0 & 45.71    &    60.1 & 61.52  &  57.0 & 62.74  \\
~ + \ourmethod{} & 77.46 & 48.67 \scriptsize{(+0.26)} &  63.84 & 46.74 \scriptsize{(+1.03)}    &    62.2 & 62.8 \scriptsize{(+1.28)} &  60.1 & \textbf{64.47} \scriptsize{(+1.73)} \\

\bottomrule
\end{tabular}
\caption{Main experimental results on the test data of CKBPv2 and SD-ATOMIC$_{20}^{20}$. F1 score is the main metric.}
\label{tab:overall_result}
\vspace{-0.1in}
\end{table*}

\vspace{-0.1in}
\subsection{Setups}
\label{sec:setup}

The following prompting methods serve as the main-task component's baselines in our experiments.

\begin{itemize}[leftmargin=*,label=$\bullet$,noitemsep,partopsep=0pt,topsep=0pt,parsep=0pt]
\item Zero-shot: standard zero-shot prompting, which directly asks the task question and ``forces'' LLMs to only return the final answer. 
\item Few-shot: standard prompting which uses exemplars to facilitate in-context learning. 
Here, we consider three variants 
which are different in the way they select exemplars. 
In detail, we use
(1) Random exemplar selection.
(2) KATE~\cite{liu-etal-2022-makes}, which chooses exemplars that are the most semantically similar to the test instance using sentence embedding models.
(3) KATE-s, 
a special version of Few-shot KATE for CSKB reasoning, in which the selected exemplars must have the same relation as the test instance.
\item Zero-shot-CoT ~\cite{kojima2022zeroshot}: the zero-shot prompting technique which uses the phrase ``Let's think step by step'' to stimulate LLMs to generate rationales before the final answer.
\item Few-shot-CoT: chain-of-thought methods which use exemplars. Here, we employ a simple method that randomly selects CoT exemplars.
\end{itemize}
Our experiments are based on two large language models ChatGPT (gpt-3.5-turbo-0301) and GPT3.5 (text-davinci-003), as they are available, stable, and two of the most capable models at the time we were conducting our experiments. 
We set temperature $T = 0$ for all experiments. 
For KATE strategy, we use the best model reported in ~\citet{liu-etal-2022-makes}.\footnote{We use the checkpoint sentence-transformers/roberta-large-nli-stsb-mean-tokens via Huggingface Transformers.}
For baselines with exemplars, all exemplars are selected from the training set provided in ~\citet{fang2023ckbpv2} regardless of labels, and the number of exemplars used in each prompt is 5 by default.

We design three prompt templates which are used to convert knowledge triples (both tested instances and exemplars) into the free-text format, then input the text to LLMs. The result of each baseline will be averaged from the results of three different prompt designs. We leave details about prompt designs in Appendix \ref{sec:prompt_design}. Nonetheless, we provide one example of how we use a prompt template to convert a knowledge triple to the free-text format as follow: \\
\textbf{Triple}: (\textit{PersonX} prepare for the competition, \texttt{xReact}, \textit{PersonX} win) \\
\textbf{Template}: Answer whether the following statement is plausible. Answer with only Yes or No: \textit{<head event>, {<human-readable template of relation>} <tail event>}. \\
$\to$ \textbf{Input Prompt}: Answer whether the following statement is plausible. Answer with only Yes or No: \textit{PersonX} prepare for the competition, {as a result, \textit{PersonX} feels} \textit{PersonX} win.

\subsection{Results and Analysis}

The experimental results are shown in Table \ref{tab:overall_result}. 
We report accuracy w.r.t all instances and (binary) F1 score w.r.t. the positive class of all baselines and our method. Following ~\citet{fang2023ckbpv2}, we choose F1 as the main metric. In the columns corresponding to F1 score, we add numbers in {\scriptsize{scriptsize}} to indicate the performance gain of \ourmethod{} over prompting methods. 

Overall, our method consistently improves\footnote{Statistically significant under one-tailed t-test with confidence level 99\%.} over all prompting methods and backbone LLMs for both benchmarks, by an average margin of 0.96\%/1.11\% and 1.64\%/1.63\% in F1 score with respect to backbone ChatGPT/GPT3.5 on CKBPv2 and SD-ATOMIC$_{20}^{20}$ respectively. Similar performance gain can be also observed in groups of non-CoT, CoT, zero-shot, few-shot baselines. Furthermore, we achieve the best result on the CKBPv2 benchmark with \ourmethod{} paired with Few-shot (Random), and also achieve the best result on the synthetic benchmark SD-ATOMIC$_{20}^{20}$ with \ourmethod{} paired with Few-shot-CoT. We note that on CKBPv2, our result is not directly comparable to results from previous works ~\cite{fang2023ckbpv2, chan2023chatgpt}, as the scale of the evaluation set, the number of exemplars used in few-shot prompting methods, and the version of LLMs of our and previous works are different.

We further analyze\footnote{Since CKBPv2 is human-curated, the analysis on the benchmark is more objective and reliable than that on our synthetic SD-ATOMIC$_{20}^{20}$.} results on CKBPv2 in-depth to point out the source of improvement of \ourmethod{} and to compare the improvement brought by our method and by main-task prompting techniques.

\begin{table*}[t]
\small
\centering
\setlength{\tabcolsep}{4.5pt} 
\renewcommand{\arraystretch}{1} 
\begin{tabular}{l|ccccc|cccc}

\toprule
Constraint (Prompt Design) & xReact & oReact & xAttr & xIntent & xNeed & xWant & xEffect & HinderedBy & Causes \\
\midrule
N/A (Zero-shot baseline) & 52.75 & 41.76 & 49.89 & 37.79 & 46.11 & 51.76 & 52.61 & 34.67 & 44.18 \\
\midrule
Typing (selected P.D.) & 54.38 & 41.76 & 51.66 & - & - & - & - & - & - \\
Typing (alternative P.D.) & 59.37 & 28.33 & 52.77 & - & - & - & - & - & - \\ 
Temporal (selected P.D.) & 45.24 & 54.27 & 49.89 & 44.88 & 48.48 & 30.41 & 42.97 & 27.76 & 31.19 \\ 
Temporal (alternative P.D.) & - & - & - & 26.13 & 17.27 & - & - & - & - \\ 
\midrule
Ambiguity (P.D. 1) & 31.15 & 41.27 & 27.01 & 13.65 & 2.9 & 41.24 & 24.4 & 3.33 & 14.85 \\ 
Ambiguity (P.D. 2) & 38.67 & 42.18 & 34.48 & 30.73 & 5.34 & 48.52 & 50.53 & 3.33 & 31.29 \\ 
\midrule
Constraint-L2M (w/o exemplars) & 45.35 & 43.16 & 43.43 & 23.92 & 24.26 & - & - & - & - \\ 
Constraint-L2M (w/ 3 exemplars) & 52.16 & 38.72 & 31.5 & 13.49 & 42.9 & - & - & - & - \\
\bottomrule

\end{tabular}
\caption{Ablation study on each constraint and prompt choice. 
P.D. in the setting names abbreviates ``prompt design''. Results in this table are F1 scores w.r.t. triples of each relation. 
The notation ``-'' indicates no change in comparing to the result of the zero-shot baseline, because we do not consider those constraint-relation pairs in the preset rules (either pre- or post-pilot-study) w.r.t. the setting.}
\label{tab:ablation}
\vspace{-0.1in}
\end{table*}

\paragraph{Where does the improvement of \ourmethod{} come from?}
We take relations \texttt{xReact, oReact, xAttr} with the Typing constraint as an example to show the effect of \ourmethod{} on GPT3.5.
Recall that when \ourmethod{} is applied, the final prediction will be the logical conjunction of predictions from the zero-shot baseline and our \ourmethod{}. 
Thus, triples that our method has an effect on are those with positive predictions from the zero-shot baseline and negative predictions from \ourmethod{}. 
As our method aims to correct False Positive (FP) predictions (and not to hurt True Positive predictions), we examine \textbf{among concerned triples}, how many cases \ourmethod{}:
\begin{itemize}[leftmargin=*,noitemsep,partopsep=0pt,topsep=0pt,parsep=0pt]
\item[1.] correctly judge a triple as it violates the constraint, and because the triple's gold label has to be 0 (except incorrect human annotation), thus helps to turn the FP prediction of the baseline to True Negative,
\item[2.] incorrectly judge a triple as it violates the constraint (in fact it does not), however, because the triple's gold label is 0, the misjudgment accidentally helps to turn the FP prediction to True Negative,
\item[3.] incorrectly judge a triple as it violates the constraint (in fact it does not), and because the triple's gold label is 1, the misjudgment undesirably turns the True Positive prediction of the baseline to False Negative.
\end{itemize}
In fact, \ourmethod{} is designated for the first category. Therefore, the more the first category happens in comparison to the second and third categories, the more reliable the improvement of \ourmethod{} is. 

We ask four external voluntary graduate NLP researchers who have at least one year of experience working on CSKBs to annotate the typing constraint status (i.e. ``satisfied'' or ``not satisfied'') of those considered triples. The Fleiss’ Kappa score of this annotation is 0.2381, and the final label is the majority vote among four annotators.
From relevant annotations and predictions, we calculate the percentage of cases falling into each mentioned category, and find that 93\% of the concerned triples fall into the first category.
Similarly, when considering other relations and other baselines, we observe the majority of the cases fall into the first category. 
That shows the valid source of improvement of \ourmethod{}.

\paragraph{Comparison of \ourmethod{} and other prompt engineering techniques}

We also compare the average effectiveness of our method with other types of prompt engineering, including 1. the use of exemplar, 2. exemplar selection, and 3. chain-of-thought.
We estimate the effectiveness (i.e. net average gain) of each prompt engineering type as the average difference of F1 score between two groups of corresponding baselines with or without the appliance of such a type.
While exemplar selection and chain-of-thought do not bring any certain benefit,
the usage of exemplars brings a remarkable improvement. In fact, exemplars help to hugely increase the recall of zero-shot baselines with GPT3.5 backbone (from 36.06\% to 66.55\% on average). However, it is deemed to the strictness of GPT3.5 in judging if a knowledge triple is commonsense, as its zero-shot baselines have a much lower recall comparing to other baselines.
Meanwhile, the improvement of \ourmethod{} is consistent over all baselines as it helps to correct False-Positive predictions. Indeed, it improves the precision of baselines and does not significantly hurt the recall. Also, thanks to the simple prompt design, our method is more efficient than the use of exemplars and CoT (Table \ref{tab:cost_estimation}).

\subsection{Ablation Study}
\label{sec:ablation}

We further conduct several additional experiments with ChatGPT on CKBPv2 to show the importance of our preset rules, constraint-checking prompt choices, and \ourmethod{}'s role as a separate module from the main-task component.

We report the result of the zero-shot baseline with each constraint and each prompt design applied in Table \ref{tab:ablation}.
We focus on F1 score of test triples of 5 relations \texttt{xReact, oReact, xAttr, xIntent, xNeed} in which \ourmethod{} has effects on according to the final preset rules, as well as of 4 extra relations \texttt{xWant, xEffect, HinderedBy, Causes}. 
We show F1 score of these relations because \texttt{xWant} and \texttt{xEffect} account for a large portion of the test set, while \texttt{HinderedBy} and \texttt{Cause} were set to be checked with temporal constraint before the pilot study.

\paragraph{Effect of Preset Rules}
In previous analyses, we show where and how \ourmethod{} improves the results of other main-task prompting methods. However, it does not mean both typing and temporal constraints are necessary. As observed in two rows, Typing (selected prompt design, P.D. for short) and Temporal (selected P.D.), of Table \ref{tab:ablation}, each constraint boosts the performance on each relation that they affect on according to the post-pilot-study preset rules. That demonstrates the importance of each selected constraint.

Similarly, we study the result regarding constraint-relation pairs which are never in or removed from the preset rules after our pilot study. 
As shown in Table \ref{tab:ablation},
for Temporal (selected P.D.) constraint, F1 score on \texttt{HinderedBy} and \texttt{Causes} are lower than the counterparts in the zero-shot baseline. Also, for other relations which are never in the preset rules of the temporal constraint, such as \texttt{xWant} and \texttt{xEffect}, the constraint often hurts the performance.
Apart from that, the Ambiguity constraint (in both prompt designs which are shown in Appendix \ref{sec:prompt_design}) also hurt the performance of all relations. We argue that ambiguity is a subjective concept, thus within the simple design philosophy of \ourmethod{}, we may not find the best way to use the constraint. Overall, the result regarding unconsidered constraint-relation pairs is consistent with our observation in the pilot study.

\paragraph{Effect of the Prompt Design}
We also ablate prompt designs of typing and temporal constraints to study the effect of constraint question design on the performance on triples of each relation.
In Table \ref{tab:ablation}, Typing (alternative P.D.) and Temporal (alternative P.D.) indicate the result w.r.t the alternative prompt design for typing and temporal constraints respectively.
In fact, the alternative prompt designs (shown in Appendix \ref{sec:prompt_design}), are formulated in a more direct way that asks if the constraint is satisfied/violated,
while our selected prompt design asks more general multiple-choice questions.
It can be seen that the alternative for typing constraint gives higher scores for \texttt{xReact} and \texttt{xAttr} but much lower scores for \texttt{oReact}. 
Also, the alternative prompt of temporal is hugely worse than the selected prompt design.
We argue that the reason could be the advantage of having more context and references when asking general multiple-choice questions than when focusing on a specific case.
That demonstrates the sensitivity of our method on constraint prompt design.




\paragraph{Effect of \ourmethod{} as an Independent Component}
\ourmethod{} is used in a plug-and-play manner, where we can get predictions independently from the main-task component. 
In this part, we study an alternative single-prompt design choice that models \ourmethod{} as an end-to-end CoT-like pipeline to directly add to the main-task prompt.
This serves as additional experiments to demonstrate the advantage of our plug-and-play design as opposed to the single-prompt counterpart.

Inspired by Least-to-Most (L2M) ~\cite{zhou2023leasttomost} which first decomposes a complex problem into a list of easier subproblems, and then sequentially solving these subproblems in different passes to LLM to reach the final answer, in this ablation, we treat constraints as easier subproblems and the main question as the hardest question which is asked ultimately.
The CoT will immediately stop and conclude that a triple is not commonsense if the triple does not satisfy the constraint. 
We name the alternative method as Constraint-L2M for simplicity. The prompt design of this baseline is shown in Appendix \ref{sec:prompt_design}.

On the one hand, results in Table \ref{tab:ablation} show that even with exemplars,
Constraint-L2M hurts the performance on all considered relations, in contrast to \ourmethod{}.
Taking a closer look into the output of Constraint-L2M (w/o exemplars) w.r.t backbone GPT3.5 and the typing constraint, we observe that 14\% of the logic step (to check if the determined tail event's type matches with the desired type, virtually equivalent to string matching) is incorrect.
Meanwhile, other combinations, though having a small error rate in the logic step, unexpectedly perform worse than the zero-shot counterparts in the main-task step. We argue that the phenomenon possibly results from the influence of patterns~\cite{turpin2023language} in constraint-checking steps' output.
That shows the advantages offered by our method's separateness from main-task prompting methods, as fusing the constraint checking step to the main-task prompt can even increase the error rate due to failures in the symbolic reasoning step or its interference on the main-task step.

On the other hand, in term of efficiency, Constraint-L2M even used without exemplars poses a much higher cost than our method (Table \ref{tab:cost_estimation}).
Thus, the plug-and-play design of our method is preferable over the single-prompt design in both aspects - effectiveness and cost.



\section{Conclusion}

In this paper, we proposed \ourmethod{}, a constraint-wise plugin, to help LLMs and prompting methods to cope with the problem of explicit relational constraint in CSKB reasoning.
Experimental results show that \ourmethod{} consistently improves over any main-task prompting technique by a significant margin, achieving SoTA performance on the benchmark.

\section*{Acknowledgement}

The authors of this paper were supported by the NSFC Fund (U20B2053) from the NSFC of China, the RIF (R6020-19 and R6021-20) and the GRF (16211520 and 16205322) from RGC of Hong Kong, and the UGC Research Matching Grants (RMGS20EG01-D, RMGS20CR11, RMGS20CR12, RMGS20EG19, RMGS20EG21, RMGS23CR05, RMGS23EG08). We also thank Samuel Cahyawijaya and Bryan Wilie from HKUST for their helpful comments.

\section*{Limitations}

This paper works on a task-specific method as well as evaluates it on two CSKB reasoning benchmarks. Further study of the proposed method on other reasoning tasks is expected to examine its broader generalizability.
Also, the process of building the first module of our method is complex and requires a certain understanding of the task and the benchmarks. How to systematically adapt this module to other tasks (e.g. automatically inducing constraints from questions), such as CSKB reasoning in the generative setting or commonsense question answering, remains to be studied. Last but not least, as the number of prompt designs, especially the set of human-readable templates, is limited, it is not 100\% guaranteed that our method will be effective for other designs of prompt of this CSKB reasoning or other reasoning tasks in general. More experiments are needed to make our claim more certain.

\section*{Ethical Considerations}

This work aims to improve the performance of LLMs on a commonsense reasoning task, which - in the case of this work - involves the use of ChatGPT (gpt-3.5-turbo-0301) and GPT3.5 (text-davinci-003). Therefore, the same risks from LLMs research are also applicable to this work~\cite{10.1145/3442188.3445922}. 

In term of computational cost, our work does not involve any training or finetuning of (large) language models. From our rough calculation which is partially shown in Table \ref{tab:cost_estimation}, the expense to conduct all experiments (including both preliminary and main experiments) in this project is around US\$200.

Last but not least, this paper works on CKBPv2 and ATOMIC$_{20}^{20}$, open-source benchmarks for the research community to study the CSKB reasoning problem under an MIT and CC-BY license respectively. This work shares the same ethical issues as the work of CKBPv2 and ATOMIC$_{20}^{20}$. Data is anonymized, thus our work does not propagate any privacy problems about any specific entities. Also, we carried out human expert annotation for analysis purposes. Since the amount of work is small, we and the annotators agree to consider it as a voluntary service.

\bibliography{custom}

\begin{thebibliography}{43}
\expandafter\ifx\csname natexlab\endcsname\relax\def\natexlab#1{#1}\fi

\bibitem[{Bender et~al.(2021)Bender, Gebru, McMillan-Major, and Shmitchell}]{10.1145/3442188.3445922}
Emily~M. Bender, Timnit Gebru, Angelina McMillan-Major, and Shmargaret Shmitchell. 2021.
\newblock \href {https://doi.org/10.1145/3442188.3445922} {On the dangers of stochastic parrots: Can language models be too big?}
\newblock In \emph{Proceedings of the 2021 ACM Conference on Fairness, Accountability, and Transparency}, FAccT '21, page 610–623, New York, NY, USA. Association for Computing Machinery.

\bibitem[{Bengio et~al.(2021)Bengio, Lecun, and Hinton}]{bengio2021deeplearning}
Yoshua Bengio, Yann Lecun, and Geoffrey Hinton. 2021.
\newblock Deep learning for {AI}.
\newblock \emph{Commun. ACM}, 64(7):58--65.

\bibitem[{Bian et~al.(2023)Bian, Han, Sun, Lin, Lu, and He}]{bian2023chatgpt}
Ning Bian, Xianpei Han, Le~Sun, Hongyu Lin, Yaojie Lu, and Ben He. 2023.
\newblock Chatgpt is a knowledgeable but inexperienced solver: An investigation of commonsense problem in large language models.
\newblock \emph{arXiv preprint arXiv:2303.16421}.

\bibitem[{Bosselut et~al.(2019)Bosselut, Rashkin, Sap, Malaviya, Celikyilmaz, and Choi}]{bosselut-etal-2019-comet}
Antoine Bosselut, Hannah Rashkin, Maarten Sap, Chaitanya Malaviya, Asli Celikyilmaz, and Yejin Choi. 2019.
\newblock \href {https://doi.org/10.18653/v1/P19-1470} {{COMET}: Commonsense transformers for automatic knowledge graph construction}.
\newblock In \emph{Proceedings of the 57th Annual Meeting of the Association for Computational Linguistics}, pages 4762--4779, Florence, Italy. Association for Computational Linguistics.

\bibitem[{Chan et~al.(2023)Chan, Cheng, Wang, Jiang, Fang, Liu, and Song}]{chan2023chatgpt}
Chunkit Chan, Jiayang Cheng, Weiqi Wang, Yuxin Jiang, Tianqing Fang, Xin Liu, and Yangqiu Song. 2023.
\newblock Chatgpt evaluation on sentence level relations: A focus on temporal, causal, and discourse relations.
\newblock \emph{ArXiv}, abs/2304.14827.

\bibitem[{Davis(2023)}]{davis2023survey}
Ernest Davis. 2023.
\newblock Benchmarks for automated commonsense reasoning: A survey.
\newblock \emph{ArXiv}, abs/2302.04752.

\bibitem[{Davison et~al.(2019)Davison, Feldman, and Rush}]{davison2019commonsense}
Joe Davison, Joshua Feldman, and Alexander~M Rush. 2019.
\newblock Commonsense knowledge mining from pretrained models.
\newblock In \emph{Proceedings of the 2019 conference on empirical methods in natural language processing and the 9th international joint conference on natural language processing (EMNLP-IJCNLP)}, pages 1173--1178.

\bibitem[{Diao et~al.(2023)Diao, Wang, Lin, and Zhang}]{diao2023active}
Shizhe Diao, Pengcheng Wang, Yong Lin, and Tong Zhang. 2023.
\newblock \href {http://arxiv.org/abs/2302.12246} {Active prompting with chain-of-thought for large language models}.

\bibitem[{Ding et~al.(2018)Ding, Wang, Wang, and Guo}]{ding-etal-2018-improving}
Boyang Ding, Quan Wang, Bin Wang, and Li~Guo. 2018.
\newblock \href {https://doi.org/10.18653/v1/P18-1011} {Improving knowledge graph embedding using simple constraints}.
\newblock In \emph{Proceedings of the 56th Annual Meeting of the Association for Computational Linguistics (Volume 1: Long Papers)}, pages 110--121, Melbourne, Australia. Association for Computational Linguistics.

\bibitem[{Fang et~al.(2023)Fang, Do, Choi, Wang, and Song}]{fang2023ckbpv2}
Tianqing Fang, Quyet~V. Do, Sehyun Choi, Weiqi Wang, and Yangqiu Song. 2023.
\newblock Ckbp v2: An expert-annotated evaluation set for commonsense knowledge base population.
\newblock \emph{ArXiv}, abs/2304.10392.

\bibitem[{Fang et~al.(2021{\natexlab{a}})Fang, Wang, Choi, Hao, Zhang, Song, and He}]{fang-etal-2021-benchmarking}
Tianqing Fang, Weiqi Wang, Sehyun Choi, Shibo Hao, Hongming Zhang, Yangqiu Song, and Bin He. 2021{\natexlab{a}}.
\newblock \href {https://doi.org/10.18653/v1/2021.emnlp-main.705} {Benchmarking commonsense knowledge base population with an effective evaluation dataset}.
\newblock In \emph{Proceedings of the 2021 Conference on Empirical Methods in Natural Language Processing, {EMNLP} 2021, Virtual Event / Punta Cana, Dominican Republic, 7-11 November, 2021}, pages 8949--8964. Association for Computational Linguistics.

\bibitem[{Fang et~al.(2021{\natexlab{b}})Fang, Zhang, Wang, Song, and He}]{DBLP:conf/www/FangZWSH21}
Tianqing Fang, Hongming Zhang, Weiqi Wang, Yangqiu Song, and Bin He. 2021{\natexlab{b}}.
\newblock \href {https://doi.org/10.1145/3442381.3450117} {{DISCOS:} bridging the gap between discourse knowledge and commonsense knowledge}.
\newblock In \emph{{WWW} '21: The Web Conference 2021, Virtual Event / Ljubljana, Slovenia, April 19-23, 2021}, pages 2648--2659. {ACM} / {IW3C2}.

\bibitem[{Hua and Zhang(2022)}]{hua-zhang-2022-system}
Wenyue Hua and Yongfeng Zhang. 2022.
\newblock \href {https://aclanthology.org/2022.findings-emnlp.42} {System 1 + system 2 = better world: Neural-symbolic chain of logic reasoning}.
\newblock In \emph{Findings of the Association for Computational Linguistics: EMNLP 2022}, pages 601--612, Abu Dhabi, United Arab Emirates. Association for Computational Linguistics.

\bibitem[{Huang and Chang(2023)}]{huang-chang-2023-towards}
Jie Huang and Kevin Chen-Chuan Chang. 2023.
\newblock \href {https://doi.org/10.18653/v1/2023.findings-acl.67} {Towards reasoning in large language models: A survey}.
\newblock In \emph{Findings of the Association for Computational Linguistics: ACL 2023}, pages 1049--1065, Toronto, Canada. Association for Computational Linguistics.

\bibitem[{Hwang et~al.(2021)Hwang, Bhagavatula, Bras, Da, Sakaguchi, Bosselut, and Choi}]{hwang2021comet}
Jena~D. Hwang, Chandra Bhagavatula, Ronan~Le Bras, Jeff Da, Keisuke Sakaguchi, Antoine Bosselut, and Yejin Choi. 2021.
\newblock \href {https://ojs.aaai.org/index.php/AAAI/article/view/16792} {(comet-) atomic 2020: On symbolic and neural commonsense knowledge graphs}.
\newblock In \emph{Thirty-Fifth {AAAI} Conference on Artificial Intelligence, {AAAI} 2021, Thirty-Third Conference on Innovative Applications of Artificial Intelligence, {IAAI} 2021, The Eleventh Symposium on Educational Advances in Artificial Intelligence, {EAAI} 2021, Virtual Event, February 2-9, 2021}, pages 6384--6392. {AAAI} Press.

\bibitem[{Ilievski et~al.(2021)Ilievski, Pujara, and Zhang}]{ilievski2021story}
Filip Ilievski, Jay Pujara, and Hanzhi Zhang. 2021.
\newblock Story generation with commonsense knowledge graphs and axioms.
\newblock In \emph{Workshop on Commonsense Reasoning and Knowledge Bases}.

\bibitem[{Kojima et~al.(2022)Kojima, Gu, Reid, Matsuo, and Iwasawa}]{kojima2022zeroshot}
Takeshi Kojima, Shixiang~(Shane) Gu, Machel Reid, Yutaka Matsuo, and Yusuke Iwasawa. 2022.
\newblock Large language models are zero-shot reasoners.
\newblock In \emph{Advances in Neural Information Processing Systems}, volume~35, pages 22199--22213.

\bibitem[{Krompa{\ss} et~al.(2015)Krompa{\ss}, Baier, and Tresp}]{denis2015typeconstraint}
Denis Krompa{\ss}, Stephan Baier, and Volker Tresp. 2015.
\newblock Type-constrained representation learning in knowledge graphs.
\newblock In \emph{The Semantic Web - ISWC 2015}, pages 640--655, Cham. Springer International Publishing.

\bibitem[{Lan et~al.(2023)Lan, He, Liu, and Zhao}]{lan2023knowledge}
Yinyu Lan, Shizhu He, Kang Liu, and Jun Zhao. 2023.
\newblock \href {http://arxiv.org/abs/2301.02781} {Knowledge reasoning via jointly modeling knowledge graphs and soft rules}.

\bibitem[{Li et~al.(2016)Li, Taheri, Tu, and Gimpel}]{li-etal-2016-commonsense}
Xiang Li, Aynaz Taheri, Lifu Tu, and Kevin Gimpel. 2016.
\newblock \href {https://doi.org/10.18653/v1/p16-1137} {Commonsense knowledge base completion}.
\newblock In \emph{Proceedings of the 54th Annual Meeting of the Association for Computational Linguistics, {ACL} 2016, August 7-12, 2016, Berlin, Germany, Volume 1: Long Papers}. The Association for Computer Linguistics.

\bibitem[{Li et~al.(2023)Li, Lin, Zhang, Fu, Chen, Lou, and Chen}]{li2023making}
Yifei Li, Zeqi Lin, Shizhuo Zhang, Qiang Fu, Bei Chen, Jian-Guang Lou, and Weizhu Chen. 2023.
\newblock \href {http://arxiv.org/abs/2206.02336} {Making large language models better reasoners with step-aware verifier}.

\bibitem[{Ling et~al.(2023)Ling, Fang, Li, Huang, Lee, Memisevic, and Su}]{ling2023deductive}
Zhan Ling, Yunhao Fang, Xuanlin Li, Zhiao Huang, Mingu Lee, Roland Memisevic, and Hao Su. 2023.
\newblock \href {http://arxiv.org/abs/2306.03872} {Deductive verification of chain-of-thought reasoning}.

\bibitem[{Liu et~al.(2022)Liu, Shen, Zhang, Dolan, Carin, and Chen}]{liu-etal-2022-makes}
Jiachang Liu, Dinghan Shen, Yizhe Zhang, Bill Dolan, Lawrence Carin, and Weizhu Chen. 2022.
\newblock \href {https://doi.org/10.18653/v1/2022.deelio-1.10} {What makes good in-context examples for {GPT}-3?}
\newblock In \emph{Proceedings of Deep Learning Inside Out (DeeLIO 2022): The 3rd Workshop on Knowledge Extraction and Integration for Deep Learning Architectures}, pages 100--114, Dublin, Ireland and Online. Association for Computational Linguistics.

\bibitem[{Malaviya et~al.(2020)Malaviya, Bhagavatula, Bosselut, and Choi}]{Malaviya2019ExploitingSA}
Chaitanya Malaviya, Chandra Bhagavatula, Antoine Bosselut, and Yejin Choi. 2020.
\newblock \href {https://ojs.aaai.org/index.php/AAAI/article/view/5684} {Commonsense knowledge base completion with structural and semantic context}.
\newblock In \emph{The Thirty-Fourth {AAAI} Conference on Artificial Intelligence, {AAAI} 2020, The Thirty-Second Innovative Applications of Artificial Intelligence Conference, {IAAI} 2020, The Tenth {AAAI} Symposium on Educational Advances in Artificial Intelligence, {EAAI} 2020, New York, NY, USA, February 7-12, 2020}, pages 2925--2933. {AAAI} Press.

\bibitem[{Mostafazadeh et~al.(2020)Mostafazadeh, Kalyanpur, Moon, Buchanan, Berkowitz, Biran, and Chu{-}Carroll}]{mostafazadeh2020glucose}
Nasrin Mostafazadeh, Aditya Kalyanpur, Lori Moon, David~W. Buchanan, Lauren Berkowitz, Or~Biran, and Jennifer Chu{-}Carroll. 2020.
\newblock \href {https://doi.org/10.18653/v1/2020.emnlp-main.370} {{GLUCOSE:} generalized and contextualized story explanations}.
\newblock In \emph{Proceedings of the 2020 Conference on Empirical Methods in Natural Language Processing, {EMNLP} 2020, Online, November 16-20, 2020}, pages 4569--4586. Association for Computational Linguistics.

\bibitem[{Pan et~al.(2023)Pan, Albalak, Wang, and Wang}]{pan2023logiclm}
Liangming Pan, Alon Albalak, Xinyi Wang, and William~Yang Wang. 2023.
\newblock \href {http://arxiv.org/abs/2305.12295} {Logic-lm: Empowering large language models with symbolic solvers for faithful logical reasoning}.

\bibitem[{Qin et~al.(2023)Qin, Zhang, Zhang, Chen, Yasunaga, and Yang}]{qin2023chatgpt}
Chengwei Qin, Aston Zhang, Zhuosheng Zhang, Jiaao Chen, Michihiro Yasunaga, and Diyi Yang. 2023.
\newblock Is chatgpt a general-purpose natural language processing task solver?
\newblock \emph{arXiv preprint arXiv:2302.06476}.

\bibitem[{Sap et~al.(2019)Sap, Bras, Allaway, Bhagavatula, Lourie, Rashkin, Roof, Smith, and Choi}]{sap2019atomic}
Maarten Sap, Ronan~Le Bras, Emily Allaway, Chandra Bhagavatula, Nicholas Lourie, Hannah Rashkin, Brendan Roof, Noah~A. Smith, and Yejin Choi. 2019.
\newblock \href {https://doi.org/10.1609/aaai.v33i01.33013027} {{ATOMIC:} an atlas of machine commonsense for if-then reasoning}.
\newblock In \emph{The Thirty-Third {AAAI} Conference on Artificial Intelligence, {AAAI} 2019, The Thirty-First Innovative Applications of Artificial Intelligence Conference, {IAAI} 2019, The Ninth {AAAI} Symposium on Educational Advances in Artificial Intelligence, {EAAI} 2019, Honolulu, Hawaii, USA, January 27 - February 1, 2019}, pages 3027--3035. {AAAI} Press.

\bibitem[{Shum et~al.(2023)Shum, Diao, and Zhang}]{shum2023automatic}
KaShun Shum, Shizhe Diao, and Tong Zhang. 2023.
\newblock \href {http://arxiv.org/abs/2302.12822} {Automatic prompt augmentation and selection with chain-of-thought from labeled data}.

\bibitem[{Speer et~al.(2017)Speer, Chin, and Havasi}]{singh2017conceptnet}
Robyn Speer, Joshua Chin, and Catherine Havasi. 2017.
\newblock \href {http://aaai.org/ocs/index.php/AAAI/AAAI17/paper/view/14972} {Conceptnet 5.5: An open multilingual graph of general knowledge}.
\newblock In \emph{Proceedings of the Thirty-First {AAAI} Conference on Artificial Intelligence, February 4-9, 2017, San Francisco, California, {USA}}, pages 4444--4451. {AAAI} Press.

\bibitem[{Turpin et~al.(2023)Turpin, Michael, Perez, and Bowman}]{turpin2023language}
Miles Turpin, Julian Michael, Ethan Perez, and Samuel~R. Bowman. 2023.
\newblock \href {http://arxiv.org/abs/2305.04388} {Language models don't always say what they think: Unfaithful explanations in chain-of-thought prompting}.

\bibitem[{Wang et~al.(2015)Wang, Wang, and Guo}]{Wang2015KnowledgeBC}
Quan Wang, Bin Wang, and Li~Guo. 2015.
\newblock Knowledge base completion using embeddings and rules.
\newblock In \emph{International Joint Conference on Artificial Intelligence}.

\bibitem[{Wang et~al.(2023{\natexlab{a}})Wang, Fang, Xu, Bo, Song, and Chen}]{wang-etal-2023-cat}
Weiqi Wang, Tianqing Fang, Baixuan Xu, Chun Yi~Louis Bo, Yangqiu Song, and Lei Chen. 2023{\natexlab{a}}.
\newblock \href {https://doi.org/10.18653/v1/2023.acl-long.733} {{CAT}: A contextualized conceptualization and instantiation framework for commonsense reasoning}.
\newblock In \emph{Proceedings of the 61st Annual Meeting of the Association for Computational Linguistics (Volume 1: Long Papers)}, pages 13111--13140, Toronto, Canada. Association for Computational Linguistics.

\bibitem[{Wang et~al.(2023{\natexlab{b}})Wang, Wei, Schuurmans, Le, Chi, Narang, Chowdhery, and Zhou}]{wang2023selfconsistency}
Xuezhi Wang, Jason Wei, Dale Schuurmans, Quoc~V Le, Ed~H. Chi, Sharan Narang, Aakanksha Chowdhery, and Denny Zhou. 2023{\natexlab{b}}.
\newblock \href {https://openreview.net/forum?id=1PL1NIMMrw} {Self-consistency improves chain of thought reasoning in language models}.
\newblock In \emph{The Eleventh International Conference on Learning Representations}.

\bibitem[{Wei et~al.(2022)Wei, Wang, Schuurmans, Bosma, brian ichter, Xia, Chi, Le, and Zhou}]{wei2022cot}
Jason Wei, Xuezhi Wang, Dale Schuurmans, Maarten Bosma, brian ichter, Fei Xia, Ed~H. Chi, Quoc~V Le, and Denny Zhou. 2022.
\newblock \href {https://openreview.net/forum?id=_VjQlMeSB_J} {Chain of thought prompting elicits reasoning in large language models}.
\newblock In \emph{Advances in Neural Information Processing Systems}.

\bibitem[{Yao et~al.(2019)Yao, Mao, and Luo}]{yao2019kg}
Liang Yao, Chengsheng Mao, and Yuan Luo. 2019.
\newblock Kg-bert: Bert for knowledge graph completion.
\newblock \emph{arXiv preprint arXiv:1909.03193}.

\bibitem[{Yu et~al.(2022)Yu, Zhang, Song, and Ng}]{yu2022cocolm}
Changlong Yu, Hongming Zhang, Yangqiu Song, and Wilfred Ng. 2022.
\newblock Cocolm: Complex commonsense enhanced language model with discourse relations.
\newblock In \emph{Findings of the Association for Computational Linguistics: ACL 2022}, pages 1175--1187.

\bibitem[{Zellers et~al.(2019)Zellers, Bisk, Farhadi, and Choi}]{zellers2019vcr}
Rowan Zellers, Yonatan Bisk, Ali Farhadi, and Yejin Choi. 2019.
\newblock From recognition to cognition: Visual commonsense reasoning.
\newblock In \emph{The IEEE Conference on Computer Vision and Pattern Recognition (CVPR)}.

\bibitem[{Zhang et~al.(2019)Zhang, Paudel, Wang, Chen, Zhu, Zhang, Bernstein, and Chen}]{zhang2019iterE}
Wen Zhang, Bibek Paudel, Liang Wang, Jiaoyan Chen, Hai Zhu, Wei Zhang, Abraham Bernstein, and Huajun Chen. 2019.
\newblock Iteratively learning embeddings and rules for knowledge graph reasoning.
\newblock In \emph{{WWW}}, pages 2366--2377. {ACM}.

\bibitem[{Zhang et~al.(2023)Zhang, Chen, Zhang, and Chen}]{zhang2023making}
Yichi Zhang, Zhuo Chen, Wen Zhang, and Huajun Chen. 2023.
\newblock Making large language models perform better in knowledge graph completion.
\newblock \emph{arXiv preprint arXiv:2310.06671}.

\bibitem[{Zhang et~al.(2022)Zhang, Zhang, Li, and Smola}]{zhang2022automatic}
Zhuosheng Zhang, Aston Zhang, Mu~Li, and Alex Smola. 2022.
\newblock \href {http://arxiv.org/abs/2210.03493} {Automatic chain of thought prompting in large language models}.

\bibitem[{Zhou et~al.(2023)Zhou, Sch{\"a}rli, Hou, Wei, Scales, Wang, Schuurmans, Cui, Bousquet, Le, and Chi}]{zhou2023leasttomost}
Denny Zhou, Nathanael Sch{\"a}rli, Le~Hou, Jason Wei, Nathan Scales, Xuezhi Wang, Dale Schuurmans, Claire Cui, Olivier Bousquet, Quoc~V Le, and Ed~H. Chi. 2023.
\newblock \href {https://openreview.net/forum?id=WZH7099tgfM} {Least-to-most prompting enables complex reasoning in large language models}.
\newblock In \emph{The Eleventh International Conference on Learning Representations}.

\bibitem[{Zhou et~al.(2021)Zhou, Gopalakrishnan, Hedayatnia, Kim, Pujara, Ren, Liu, and Hakkani-Tur}]{zhou2021commonsense}
Pei Zhou, Karthik Gopalakrishnan, Behnam Hedayatnia, Seokhwan Kim, Jay Pujara, Xiang Ren, Yang Liu, and Dilek Hakkani-Tur. 2021.
\newblock Commonsense-focused dialogues for response generation: An empirical study.
\newblock In \emph{Proceedings of the 22nd Annual Meeting of the Special Interest Group on Discourse and Dialogue}, pages 121--132.

\end{thebibliography}
\bibliographystyle{acl_natbib}

\appendix
\newpage

\section{Experiments}
In this section, we provide details of how to form test benchmarks, the pilot study, additional analyses of results, and baselines.

\subsection{Benchmarks}
\label{sec:benchmark}

CKBPv2 originally consists of approximately 1k development instances and 4k test instances. To reduce the computational cost while keeping the same data distribution, we use stratified sampling to down-scale the test split of the benchmark by a factor of 4, hence forming a test set of 979 instances. The down-sampled test set includes 208 instances with label 1 (which means they are commonsense or ``positive''), thus, the ratio of the number of commonsense/not commonsense instances remains approximately 1/4.
In fact, results of the human-performance baseline and supervised-learning baseline in Table \ref{tab:dataset_previous_result} suggest that the down-scaled test set is representative of the whole test set.

Meanwhile, SD-ATOMIC$_{20}^{20}$ is curated from the test set of ATOMIC$_{20}^{20}$ as follow. First, we randomly select 1000 distinct head events, and then for each head event, we select one triple (i.e. instance). Since our final preset rules concern 5 relations \texttt{xReact, oReact, xAttr, xIntent, xNeed}, the selected triples are also of these 5 relations only. Finally, we apply a data-manipulation method on the collection of 1000 triples to sample negative instances for SD-ATOMIC$_{20}^{20}$, in which we randomly select (1) 250 triples and change the relation, and (2) 250 triples (h,r,t) and change the tail event so that obtained triples do not exist in the ATOMIC$_{20}^{20}$'s test set.
By assumption, these 500 instances are not commonsense and then assigned the label 0, while the rest 500 instances which remain intact are commonsense and assigned the label 1.

\begin{table}[t]
\small
\centering
\setlength{\tabcolsep}{4.5pt} 
\renewcommand{\arraystretch}{1} 
\begin{tabular}{l|llll}
\toprule
Method & Acc. & Pre. & Rec. & F1 \\ 
\midrule
Random ($p = 0.5$) & 50.00 & 20.00 & 50.00 & 28.57 \\ 
Human \textit{(full set)} & - & - & - & 91.50 \\ 
Prior best \textit{(full set)} & - & - & - & 46.63 \\ 
Human & 96.38 & 94.37 & 88.22 & 91.17 \\ 
Prior best & 60.54 & 32.08 & 76.76 & 45.25 \\ 

\bottomrule
\end{tabular}
\caption{Random, Human, and previous best Supervised Learning baselines' performance as a lower bound, upper bound, and a competitive baseline to compare with LLM prompting methods. Here, Acc., Pre., Rec. respectively mean Accuracy, Precision, and Recall. The random baseline follows the Bernoulli distribution with probability $p$ is $0.5$. Results of baselines with suffix \textit{(full set)} are results on the whole test set of CKBPv2~\cite{fang2023ckbpv2}. For the last two rows, we use the available human annotation of CKBPv2 and rebuild the best baseline based on its description in~\citet{fang2023ckbpv2} to compute these statistics. Results in this table suggest that the down-scaled test set is representative of the whole test set.}
\label{tab:dataset_previous_result}
\end{table}

\subsection{Pilot Study}
\label{sec:pilot}
We sampled 102 instances from the dev split of CKBPv2 in a relation-wise stratified manner to form a small dataset for the pilot study. The prompt design used in this pilot study consists of zero-shot template design 3 (Table \ref{tab:main_task_prompt}), constraint template design 1 for Typing and Temporal and template design 2 for Ambiguity (Table \ref{tab:constraint_prompt}). The result is shown in Table \ref{tab:pilot}, with a similar organization as Table \ref{tab:ablation}.

\begin{table*}[t]
\small
\centering
\setlength{\tabcolsep}{4.5pt} 
\renewcommand{\arraystretch}{1} 
\begin{tabular}{l|c|ccccc|cccc}

\toprule
Constraint & All & xReact & oReact & xAttr & xIntent & xNeed & xWant & xEffect & HinderedBy & Causes \\
\midrule
N/A (Baseline) & 61.29 & 66.67 & 100.00 & 75.00 & 50.00 & 50.00 & 66.67 & 66.67 & 100.00 & 0.00 \\
Typing & 72.73 & 62.30 & 100.00 & 75.00 & - & - & - & - & - & - \\
Temporal & 64.41 & - & - & - & 100.00 & 57.14 & - & - & 100.00 & 0.00 \\
Ambiguity & 29.27 & 25.00 & 0.00 & 57.14 & 66.67 & 0.00 & 57.14 & 0.00 & 0.00 & 0.00 \\
\bottomrule

\end{tabular}
\caption{Result in the pilot study. The result is presented in a similar organization as Table \ref{tab:ablation}}
\label{tab:pilot}
\vspace{-0.1in}
\end{table*}

We observed no effect of the Ambiguity constraint, thus we dropped that constraint. Furthermore, as we observed no effect of the Temporal constraint on samples of relations \texttt{HinderedBy} and \texttt{Causes}, we decided to remove these constraint-relation pairs. Nonetheless, while there is no effect of the Typing constraint on samples of relations \texttt{oReact} and \texttt{xAttr}, we still kept these constraint-relation pairs because the readable templates of the two relations do not adequately reflect their Typing constraint.

\subsection{Analysis}
\label{sec:analysis}

As observed in Table \ref{tab:overall_result}, two groups of non-CoT and CoT baselines have the results w.r.t ChatGPT and GPT3.5 showing different patterns. 
While non-CoT baselines with backbone ChatGPT do not benefit much from or even suffer from a performance decrease due to exemplars, the non-CoT baselines with GPT3.5 and all CoT baselines hugely benefit from in-context exemplars. In CKBPv2, baselines with in-context exemplars are 6\% to 10\% better than their corresponding zero-shot counterparts.

\begin{table}[t]
\small
\centering
\setlength{\tabcolsep}{4.5pt} 
\renewcommand{\arraystretch}{1.1} 
\begin{tabular}{l|ll|ll}
\toprule
\multirow{2}{*}{Method} & \multicolumn{2}{c|}{ChatGPT} & \multicolumn{2}{c}{GPT3.5} \\
\cline{2-5} & Acc & F1 & Acc & F1 \\

\midrule
Active-CoT & 74.67 & 49.29  &  72.28 & 50.39 \\
Automate-CoT & 76.0 & 50.82  &  76.68 & 50.31 \\

\bottomrule
\end{tabular}
\caption{Results of extra exemplar-optimization baselines on CKBPv2.}
\label{tab:extra_exemplar_optimization}
\vspace{-0.1in}
\end{table}

For CKBPv2, we try two exemplar optimization methods - Active-CoT ~\cite{diao2023active} and Automate-CoT ~\cite{shum2023automatic} which respectively use uncertainty-based active learning and rational chains optimization for exemplar selection (Table ~\ref{tab:extra_exemplar_optimization}). We notice that exemplar optimization becomes more important for CoT baselines, as the optimization gives a significant gain on ChatGPT and a large improvement on GPT3.5.
Also, CoT baselines generally achieve higher score than non-CoT baselines, which is often observed in other benchmarks. 

We further explore the dependence of overall baseline performance on our 3 seed prompt designs.
The average precision, recall, F1 score over all baselines w.r.t to each prompt design is reported in Table \ref{tab:seed_prompt_analysis}.
There is a gap between the third seed prompt design and other two seed prompt designs, however, the gap is not significantly large.
Therefore, we conclude that there is no significant dependence of baseline performance on seed prompt designs. 

\begin{table}[t]
\small
\centering
\setlength{\tabcolsep}{4.5pt} 
\renewcommand{\arraystretch}{1} 
\begin{tabular}{l|ccc}

\toprule
Prompt design & 1 & 2 & 3 \\
\midrule
Precision & 39.85 & 39.76 & 39.63 \\
Recall & 59.95 & 59.31 & 60.37 \\
F1 score & 46.38 & 46.32 & 47.15 \\
\bottomrule

\end{tabular}
\caption{Average Precision, Recall, and F1 score over all baselines of each seed prompt design. There is no significant dependence of baseline performance on seed prompt designs.}
\label{tab:seed_prompt_analysis}
\vspace{-0.1in}
\end{table}

Last but not least, we examine to what extent LLMs fail to handle the explicit constraints. We focus on a specific context, which considers the prediction of Few-shot-CoT baseline (with ChatGPT backbone and the third prompt design) and the \texttt{xReact} relation. As the Few-shot-CoT baseline works on the main-task question of whether a triple is commonsense, its prediction is not equivalent to the prediction of whether the triple satisfies the constraint. Only its ``Yes'' prediction implies a ``Yes'' prediction of constraint satisfaction. Thus, we estimate the failure rate of the Few-shot-CoT baseline based on triples with positive predictions.
Among those triples, 43\% of the triples do not satisfy the typing constraint, but the baseline implicitly predicts them as satisfied. That supports our claim that LLMs and advanced prompting techniques become less effective in handling explicit constraints in CSKB reasoning.

\subsection{Cost Estimation}

In Table \ref{tab:cost_estimation}, we estimate the total number of words processed for each instance in each baseline as well as the overhead cost of using additional prompting engineering techniques.
We ignore the cost of exemplar optimization, as the process is done at most once per baseline and independent of the size of the test set.
As such, here we treat Few-shot (KATE(-s)) the same as Few-shot (Random), and treat Active-CoT and Automate-CoT the same as Few-shot-CoT.
Also, since \ourmethod{}'s constraint design only involves the head and tail events of test triples, which are irrelevant to seed prompt designs for the main-task component, we only need to run \ourmethod{} once then apply for all baselines and for all seed prompt designs.
Overall, it shows the efficiency of \ourmethod{} over other types of prompt engineering.

\begin{table}[t]
\small
\centering
\setlength{\tabcolsep}{6pt} 
\renewcommand{\arraystretch}{1.1} 
\begin{tabular}{l|c}

\toprule
Method & Words \\
\midrule
\multicolumn{2}{l}{Representative baselines} \\
\midrule
Zero-shot & 28 \\
Few-shot & 120 \\
Zero-shot-CoT & 68 \\
Few-shot-CoT & 321 \\
\midrule
\multicolumn{2}{l}{Type of prompt engineering} \\
\midrule
Using exemplars & 172 \\
Using CoT & 120 \\
\ourmethod{} & 72 \\
Constraint-L2M (w/o exemplar) & 254 \\
\bottomrule

\end{tabular}
\caption{Per-instance estimated costs for baselines and additional costs for each type of prompt engineering, including ``Using exemplars'', ``Using CoT'', and \ourmethod{}.
``Words'' indicates the average number of words in the input prompt and generated which are both charged by OpenAI, which are proportional to the actual costs.
For types of prompt engineering other than \ourmethod{}, we take the average gap between two groups of baselines w.r.t the type. Here, we generously assume that all constraints of \ourmethod{} are run for every instance, instead of only instances of concerned relations as following the preset rules.}
\label{tab:cost_estimation}
\vspace{-0.1in}
\end{table}

\subsection{Why is ConstraintChecker not extended to intervene False Negatives?}

Intervening on False Negatives is equivalent to using constraints satisfaction to ``convince'' the main-task component that a triple is correct. However, constraint satisfaction is not adequate to justify if the triple is correct, as we also need to consider its overall semantic. 

A real example from the development split of CKBPv2 is (\textit{PersonX} go to sleep on hollow, \texttt{xReact}, \textit{PersonX} be tired). Clearly, ``\textit{PersonX} be tired'' expresses a mental state, which means the triple satisfies the typing constraint corresponding to \texttt{xReact} that the tail event has to express a mental state. However, the phrase ``sleep on hollow'' is ambiguous, and even if we ignore the words ``on hollow'', it’s unlikely that ``\textit{PersonX} be tired'' is a result of ``\textit{PersonX} go to sleep''. That means the triple is not common sense.

\section{Supplementary Materials}

\subsection{Taxonomy of CSKB relations}

The taxonomy of CSKB relations are aggregrated from prior works ~\cite{sap2019atomic, hwang2021comet} and demonstrated in Table \ref{tab:relation_taxonomy}.

\begin{table*}[t]
\small
\centering
\setlength{\tabcolsep}{4.5pt} 
\renewcommand{\arraystretch}{1} 
\begin{tabular}{l|l}

\toprule
Type & Relations \\
\midrule
\multicolumn{2}{c}{ATOMIC~\cite{sap2019atomic}} \\
\midrule
Event & \texttt{xNeed, xEffect, xWant, oEffect, oWant} \\
Mental state & \texttt{xIntent, xReact, oReact} \\
Persona & \texttt{xAttr} \\
\midrule
\multicolumn{2}{c}{ATOMIC$^{20}_{20}$~\cite{hwang2021comet}} \\
\midrule
Physical-Entity & \texttt{ObjectUse, AtLocation, MadeUpOf, HasProperty, CapableOf, Desires, Not Desires} \\
Event-Centered & \texttt{IsAfter, HasSubEvent, IsBefore, HinderedBy, Causes, xReason, isFilledBy} \\
Social-Interaction & \texttt{xNeed, xAttr, xEffect, xReact, xWant, xIntent, oEffect, oReact, oWant} \\
\bottomrule

\end{tabular}
\caption{Taxonomy of CSKB relations, cited from previous works ~\cite{sap2019atomic, hwang2021comet}. In the context of this work, we are only interested in 15 relations included in CKBPv2.}
\label{tab:relation_taxonomy}
\vspace{-0.1in}
\end{table*}

\subsection{Prompt design}
\label{sec:prompt_design}

For each triple (head event, relation, tail event), we convert the triple into a free-text format (so-called \textit{assertion}) using human-readable templates.
Along with the original set of templates in ~\citet{hwang2021comet}, we also design and experiment with another set of templates to study the correlation between human-readable template design and the result.
Likewise, we take the direct question-answering prompt (so-called \textit{main question}) design from ~\citet{fang2023ckbpv2} and self-curate another one. The two sets of human-readable templates and two main question designs are shown in the following tables.

An input prompt to LLMs consists of two main parts, the main question and the assertion.
We select three combinations of human-readable templates of relations and main question designs as \textit{seed prompt designs}, from which each baseline will adapt to get its three prompt designs (if necessary).
The result of each baseline will be averaged from the results of three different prompt designs.

\begin{table*}[t]
\small
\centering
\setlength{\tabcolsep}{4.5pt} 
\renewcommand{\arraystretch}{1} 
\begin{tabular}{l|l}

\toprule
Relation & Template \\
\midrule
xWant & as a result, PersonX wants to \\
oWant & as a result, PersonY or others want to \\
xEffect & as a result, PersonX will \\
oEffect & as a result, PersonY or others will \\
xReact & as a result, PersonX feels \\
oReact & as a result, PersonY or others feel \\
xAttr & PersonX is seen as \\
xIntent & because PersonX wanted \\
xNeed & but before, PersonX needed \\
Causes & causes \\
xReason & because \\
isBefore & happens before \\
isAfter & happens after \\
HinderedBy & can be hindered by \\
HasSubEvent & includes the event or action \\
\bottomrule

\end{tabular}
\caption{Readable templates from ~\citet{hwang2021comet} (denoted as H-template), concerning 15 relations which comprise CKBPv2. When these templates are used in the main-task component, the head and tail events will be respectively prepended and appended to the templates.}
\label{tab:readable_template}
\vspace{-0.1in}
\end{table*}

\begin{table*}[t]
\small
\centering
\setlength{\tabcolsep}{4.5pt} 
\renewcommand{\arraystretch}{1} 
\begin{tabular}{l|l}

\toprule
Relation & Template \\
\midrule
xWant & \textit{<h>}, thus, \textit{<t>} \\
oWant & \textit{<h>}, thus, \textit{<t>} \\
xEffect & \textit{<h>}, thus as an result, \textit{<t>} \\
oEffect & \textit{<h>}, thus as an result, \textit{<t>} \\
xReact & \textit{<h>}, thus as a result on PersonX's emotion, \textit{<t>} \\
oReact & \textit{<h>}, thus as a result on PersonY's emotion, \textit{<t>} \\
xAttr & \textit{<h>}, thus it can be seen about PersonX's attribute that \textit{<t>} \\
xIntent & \textit{<h>}, thus it can be seen about PersonX's intention that \textit{<t>} \\
xNeed & The event \textit{<h>} will not happen unless \textit{<t>} \\
Causes & Because \textit{<h>}, \textit{<t>} \\
xReason & \textit{<h>}, because \textit{<t>} \\
isBefore & After \textit{<h>}, \textit{<t>} \\
isAfter & Before \textit{<h>}, \textit{<t>} \\
HinderedBy & The event \textit{<h>} will not happen, if \textit{<t>} \\
HasSubEvent & The event \textit{<h>} includes the event/action that \textit{<t>} \\
\bottomrule

\end{tabular}
\caption{Self-curated readable templates (denoted as S-template) for 15 relations in CKBPv2. \textit{<h>} and \textit{<t>} denote the head and tail event respectively.}
\vspace{-0.1in}
\end{table*}

\begin{table*}[t]
\small
\centering
\setlength{\tabcolsep}{4.5pt} 
\renewcommand{\arraystretch}{1} 
\resizebox{\linewidth}{!}{%
\begin{tabular}{p{0.1\linewidth} | p{0.9\linewidth}}

\toprule
Constraint & Prompt Designs \\

\midrule
Typing & \begin{tabular}[c]{p{0.95\linewidth}}
\textbf{Design 1 (selected):}\\
Which aspect (among three options 1. event/activity, 2. persona, 3. mental state) of the subject does the clause \textit{<t>} express? Answer the choice only.\\
\\
\textbf{Design 2 (alternative):}\\
Determine if clause \textit{<t>} expresses an event or activity of the subject. Answer ``Yes'' or ``No'' only.
\end{tabular} \\

\midrule
Temporal & \begin{tabular}[c]{p{0.95\linewidth}}
\textbf{Design 1 (selected):}\\
Which one of the following two statements is more plausible: \\
0. \textit{<t>} before \textit{<h>}, \\
1. \textit{<t>} after \textit{<h>}. \\
Answer 0 or 1 only.\\
\\
\textbf{Design 2 (alternative):}\\
Judge if the event \textit{<t>} likely occurs after the event \textit{<h>}. Answer ``Yes'' or ``No'' only.
\end{tabular} \\

\midrule
Ambiguity & \begin{tabular}[c]{p{0.95\linewidth}}
\textbf{Design 1:}\\
Which one of the following two statements make more sense: \\
0. Two clauses \textit{<h>} and \textit{<t>} all have clear meaning. \\
1. One of two following clauses \textit{<h>} and \textit{<t>} has ambiguous meaning. \\
Answer 0 or 1 only.\\
\\
\textbf{Design 2:}\\
Judge if the meaning of the clauses \textit{<h>} and \textit{<t>} are all clear. Answer 'Yes' or 'No' only.
\end{tabular} \\
\bottomrule

\end{tabular}
}
\caption{Constraint prompt designs for typing, temporal, and ambiguity constraints. \textit{<h>} and \textit{<t>} denote the head and tail event respectively.}
\label{tab:constraint_prompt}
\vspace{-0.1in}
\end{table*}

\begin{table*}[t]
\small
\centering
\setlength{\tabcolsep}{4.5pt} 
\renewcommand{\arraystretch}{1} 
\resizebox{\linewidth}{!}{%
\begin{tabular}{p{0.1\linewidth} | p{0.9\linewidth}}

\toprule
Baselines & Prompt Designs \\

\midrule
Zero-shot & \begin{tabular}[c]{p{0.95\linewidth}}
\textbf{Design 1:}\\
Answer whether the following statement is plausible. Answer with only Yes or No: <free-text H-template format of the test triple> \\
\\
\textbf{Design 2:}\\
Answer whether the following statement is plausible. Answer with only Yes or No: <free-text S-template format of the test triple> \\
\\
\textbf{Design 3:}\\
Judge the following statement if it's likely to occur, only answer True or False: <free-text S-template format of the test triple> \\
\end{tabular} \\

\midrule
Few-shot & \begin{tabular}[c]{p{0.95\linewidth}}
\textbf{Design 1:}\\
Answer whether the following statement is plausible. Answer with only Yes or No:
Statement: If PersonX push PersonY back, as a result, PersonY or others will, PeopleX step back from PersonX\\
Answer: Yes\\
Statement: If PersonX regain PersonY 's composure, can be hindered by, PersonY be disown personx\\
Answer: Yes\\
Statement: If PersonX be nowhere, can be hindered by, PersonX friend will not keep PersonY\\
Answer: Yes\\
Statement: If PersonX chase PersonZ away, as a result, PersonX will, PersonY lose friend\\
Answer: Yes\\
Statement: If PersonX wave PersonY away, as a result, PersonX will, PersonY roll PersonZ eye\\
Answer: Yes\\
Statement: <free-text H-template format of the test triple>\\
Answer: \\
\\
\textbf{Design 2:}\\
Answer whether the following statement is plausible. Answer with only Yes or No:\\
Statement: PersonX stay away from PersonY, thus as an result, PersonX call out to PersonX\\
Answer: No\\
Statement: PersonX help the PersonY, thus as an result, PersonX be rebuff by PersonY\\
Answer: Yes\\
Statement: PersonX turn down that, thus it can be seen about PersonX's attribute that PersonX get PersonY into trouble\\
Answer: No\\
Statement: PersonX be real, thus as an result, PersonY argue PersonZ about it\\
Answer: Yes\\
Statement: PersonX challenge PersonZ 's friend, thus, PersonY want PersonY not let\\
Answer: Yes\\
Statement: <free-text S-template format of the test triple> \\
Answer: \\ 
\\
\textbf{Design 3:}\\
Judge the following statement if it's likely to occur, only answer True or False:\\
Statement: PersonX get PersonX thing together, thus it can be seen about PersonX's attribute that PersonX be helpful \\
Answer: True \\
Statement: PersonX invite PersonY to lunch, thus, PersonY want PersonX be a leader \\
Answer: False \\
Statement: PersonX catch, thus as a result on PersonY's emotion, PersonY feel PersonY be fluster \\
Answer: True \\
Statement: PersonX break PersonX glass, thus as a result on PersonX's emotion, PersonX feel PersonX be ashamed \\
Answer: True \\
Statement: The event PersonX need to set plan will not happen unless PersonX know about it \\
Answer: False \\
Statement: <free-text S-template format of the test triple> \\
Answer: \\ 
\end{tabular} \\

\midrule
Zero-shot-CoT & \begin{tabular}[c]{p{0.95\linewidth}}
\textbf{Design 1:}\\
Answer whether the statement <free-text H-template format of the test triple> is plausible. Let's think step by step, then conclude by answering True or False. \\
\textbf{Design 2:}\\
Answer whether the statement <free-text S-template format of the test triple> is plausible. Let's think step by step, then conclude by answering True or False. \\
\end{tabular} \\

\bottomrule
\end{tabular}
}
\caption{Prompt designs of baselines. H-template and S-template denote two sets of readable templates, from \citet{hwang2021comet} and self-curated. For baselines which select exemplars randomly or based on the test instance, the exemplars will change according to each test instance, rather than remaining the same as what are shown here.}
\label{tab:main_task_prompt}
\vspace{-0.1in}
\end{table*}

\begin{table*}[t]
\small
\centering
\setlength{\tabcolsep}{4.5pt} 
\renewcommand{\arraystretch}{1} 
\resizebox{\linewidth}{!}{%
\begin{tabular}{p{0.1\linewidth} | p{0.9\linewidth}}

\toprule
Baselines & Prompt Designs \\

\midrule
Zero-shot-CoT & \begin{tabular}[c]{p{0.95\linewidth}}
\textbf{Design 3:}\\
Judge the statement <free-text S-template format of the test triple> if it's likely to occur. Let's think step by step, then conclude by answering True or False.
\end{tabular} \\

\midrule
Few-shot-CoT & \begin{tabular}[c]{p{0.95\linewidth}}
\textbf{Design 1:}\\
Q: Answer whether the following statement is plausible: PersonX find PersonY happens before PersonX wake up on ground.\\
A: \hl{Let's think step by step. If PersonX just wake up on the ground, that means PersonX was sleeping before that. Thus, PersonX cannot be conscious to find another person.} Thus, the statement is not likely to occur.\\
\\
Q: Answer whether the following statement is plausible: PeopleX deserve happiness, as a result, PersonX feels, PersonX reach out to PeopleX.\\
A: \hl{Let's think step by step. The events 'PeopleX deserve happiness' and 'PersonX reach out to PeopleX' are likely irrevelant. Also, the clause 'PersonX reach out to PeopleX' does not describe what PersonX feels.} Thus, the statement is not likely to occur.\\
\\
Q: Answer whether the following statement is plausible: PersonX have a sheet, but before, PersonX needed PersonX meet PersonY requirement.\\
A: \hl{Let's think step by step. If PersonX doesn't meet PersonY requirement, PersonX likely doesn't get the reward from PersonY. However, in this case, it's not clear whether 'a sheet' refer to PersonY's reward or not.} Thus, the statement is not likely to occur.\\
\\
Q: Answer whether the following statement is plausible: PersonX occupy PersonY position, as a result, PersonX wants to PersonY want to aid in position.\\
A: \hl{Let's think step by step. When PersonX occupy PersonY position, it means PersonY already worked at this position and has experience to do the job. Therefore, it's likely that PersonX want PersonY to aid PerosonX when PersonX is in that job position.} Thus, the statement is likely to occur.\\
\\
Q: Answer whether the following statement is plausible: PersonX see that, as a result, PersonX will PersonX want a pet.\\
A: \hl{Let's think step by step. In this context, we can refer the word 'that' as some activity where people play with their pet. Therefore, it stimulates PersonX's desire to have a pet.} Thus, the statement is likely to occur.\\
\\
Q: Answer whether the following statement is plausible: <free-text H-template format of the test triple>.\\
A: 
\end{tabular} \\

\midrule
Constraint-L2M & \begin{tabular}[c]{p{0.95\linewidth}}
\textbf{Design 1:}\\
For each statement below, please answer several questions to reach the final conclusion if the statement is commonsense. \\
Whenever your answer of a question is No, please conclude that the statement is not commonsense. Otherwise, please conclude that the statement is commonsense. \\
 \\
Statement: If PersonX prevent PersonY, as a result, PersonX feels, PersonX never reach out to anyone \\
Q: Which aspect (among three options 1. event/activity, 2. persona, 3. mental state) of the subject does the clause 'PersonX never reach out to anyone' express. Answer the choice only. \\
A: 1. event/activity \\
Q: Is the above answer different from option 1. event/activity? \\
A: No. Thus, the statement is not commonsense \\
 \\
Statement: If PersonX go to sleep on hollow, as a result, PersonX feels, PersonX be tired \\
Q: Which aspect (among three options 1. event/activity, 2. persona, 3. mental state) of the subject does the clause 'PersonX feel PersonX be tired' express. Answer the choice only. \\
A: 3. mental state \\
Q: Is the above answer different from option 1. event/activity? \\
A: Yes \\
Q: Is the statement "If PersonX go to sleep on hollow, as a result, PersonX feels, PersonX be tired" plausible? \\
A: No. Thus, the statement is not commonsense \\
 \\
Statement: If PersonX eat the sub, as a result, PersonX feels, PersonX be full \\
Q: Which aspect (among three options 1. event/activity, 2. persona, 3. mental state) of the subject does the clause 'PersonX feel PersonX be full' express. Answer the choice only. \\
A: 3. mental state \\
Q: Is the above answer different from option 1. event/activity? \\
A: Yes \\
Q: Is the statement "If PersonX eat the sub, as a result, PersonX feels, PersonX be full" plausible? \\
A: Yes. Thus, the statement is commonsense \\
\end{tabular} \\

\bottomrule

\end{tabular}
}
\caption{(Cont.) Prompt designs of baselines. H-template and S-template denote two sets of readable templates, from \citet{hwang2021comet} and self-curated. For baselines which select exemplars randomly or based on the test instance, the exemplars will change according to each test instance, rather than remaining the same as what are shown here. For concision, we only show \textbf{Design 1} for Few-shot-CoT and Constraint-L2M.}
\vspace{-0.1in}
\end{table*}

\end{document}